\documentclass{article} 
\usepackage{iclr2021_conference,times}
\usepackage{xcolor}
\usepackage{subcaption}

\usepackage{amsmath,amsfonts,bm}

\def\eqref#1{equation~\ref{#1}}

\def\1{\bm{1}}

\DeclareMathAlphabet{\mathsfit}{\encodingdefault}{\sfdefault}{m}{sl}
\SetMathAlphabet{\mathsfit}{bold}{\encodingdefault}{\sfdefault}{bx}{n}

\usepackage{hyperref}
\usepackage{url}
\usepackage{multirow}
\usepackage{tabularx}
\usepackage{longtable}
\usepackage{booktabs}
\usepackage{graphicx}
\usepackage[boxed,linesnumbered,ruled]{algorithm2e}
\usepackage[breakable,skins]{tcolorbox}

\title{ChatEval: Towards better LLM-based evaluators through multi-agent debate}

\author{Chi-Min Chan, Weize Chen, Yusheng Su, Jianxuan Yu, Zhiyuan Liu\thanks{Corresponding author. Email: \texttt{liuzy@tsinghua.edu.cn}} \\
Department of Computer Science and Technology\\
Tsinghua University\\
\texttt{zorowin123@gmail.com} \\
\And
Jie Fu, Wei Xue \\
Hong Kong University of Science and Technology \\
\And
Shanghang Zhang \\
Peking University \\
}

\iclrfinalcopy 
\begin{document}

\maketitle

\begin{abstract}
Text evaluation has historically posed significant challenges, often demanding substantial labor and time cost. With the emergence of large language models (LLMs), researchers have explored LLMs' potential as alternatives for human evaluation. While these single-agent-based approaches show promise, experimental results suggest that further advancements are needed to bridge the gap between their current effectiveness and human-level evaluation quality. 
Recognizing that best practices of human evaluation processes often involve multiple human annotators collaborating in the evaluation, we resort to a multi-agent debate framework, moving beyond single-agent prompting strategies.
The multi-agent-based approach enables a group of LLMs to synergize with an array of intelligent counterparts, harnessing their distinct capabilities and expertise to enhance efficiency and effectiveness in handling intricate tasks. In this paper, we construct a multi-agent referee team called \textbf{ChatEval} to autonomously discuss and evaluate the quality of generated responses from different models on open-ended questions and traditional natural language generation (NLG) tasks. We derive insights and lessons from practical scenarios where humans instigate group discussions for brainstorming and propose different communication strategies within ChatEval. Our experiments on two benchmark tasks illustrate that ChatEval delivers superior accuracy and correlation in alignment with human assessment. Furthermore, we find that the diverse role prompts (different personas) are essential in the multi-agent debate process; that is, utilizing the same role description in the prompt can lead to a degradation in performance. Our qualitative analysis also shows that ChatEval transcends mere textual scoring, offering a human-mimicking evaluation process for reliable assessments.
Our code is available at \url{https://github.com/chanchimin/ChatEval}.

\end{abstract}

\section{Introduction}
\label{sec:introduction}

Evaluating the quality of text generated by language models or written by humans has long been a challenging endeavor, consistently garnering substantial attention~\citep{celikyilmazEvaluationText2021a}. Traditional methodologies predominantly rely on human annotation of texts~\citep{callison2009fast}, an approach considered overly demanding in terms of time and cost. Automatic evaluation metrics based on n-grams, such as Rouge~\citep{lin2004rouge}, BLEU~\citep{papineni2002bleu}, and METEOR~\citep{banerjee2005meteor}, have been proposed to tackle this issue~\citep{kondrak2005n}. However, these methods have been shown to exhibit a relatively weak correlation with human judgments, particularly in the context of tasks involving open-ended generation or requiring domain-specific expertise~\citep{novikova-etal-2017-need}.

Recent advancements in the field of natural language processing have led to the emergence of billion-parameter scale LLMs, such as GPT-3~\citep{brown2020language}. These LLMs have demonstrated remarkable capabilities across diverse downstream tasks, presenting new opportunities for text quality evaluation using such models. Moreover, various training paradigms have been proposed to endow LLMs with the ability to accomplish tasks in a zero-shot manner and better adhere to human-provided instructions~\citep{ouyang2022training, sanh2021multitask, wei2021finetuned}. These advancements facilitate the prompting of LLMs to evaluate generated text, effectively simulating human evaluators in the assessment process.

In view of the impressive text understanding and instruction-following capabilities of recent LLMs, a body of literature~\citep{liu2023gpteval, chiang2023can, gao2023human, shen2023large} has adopted LLM as an evaluator to assess the quality of responses to open-ended questions or traditional NLG tasks, including dialogue response generation and summarization. This methodology is dubbed LLM-as-a-judge~\citep{zheng2023judging}. Findings from these researches indicate that LLM can mimic human behavior and provide evaluations that correspond with human judgments, revealing a potentially scalable and transparent alternative to costly and laborious human evaluations.

While a \textit{single} powerful LLM can already tackle various missions, emerging studies suggest that \textit{multiple} LLMs can further improve one another through debate and cooperation~\citep{li2023camel, liang2023encouraging}. By incorporating multiple LLMs into an integrated group and designing specific interaction mechanisms, different LLMs can engage in proposing and deliberating unique responses and thought processes across several rounds. This approach leads to enhanced factuality of generated responses~\citep{du2023improving} and improvement in the completion of arduous tasks~\citep{li2023camel, qian2023communicative}. Furthermore, the multi-agent group also addresses and mitigates the Degeneration-of-Thought (DOT) problem~\citep{liang2023encouraging}.

In the human evaluation processes, relying on a single perspective can introduce bias and instability in the results~\citep{karpinska2021perils}. Recognizing this, best practices often involve multiple human annotators collaborating in the evaluation~\citep{van2019best}. Drawing inspiration from this collaborative and iterative human evaluation approach, we propose ChatEval, a system that enables each agent to employ varied communication strategies in collaborative discussion, working towards formulating final judgments. Furthermore, to enrich the evaluation dynamics, every agent within ChatEval is endowed with a unique persona. This deliberate design ensures that each agent focuses on distinct perspectives or brings specific expertise to the table. By doing so, the collective evaluation benefits from a more comprehensive lens, capturing nuances and subtleties that a single perspective might overlook. We derive this idea primarily from the insight of \textit{'There are a thousand Hamlets in a thousand people's eyes'}, meaning that every person has their unique interpretation or perspective, especially applicable to text evaluation. Indeed, these divergent perspectives shape the comprehensive and multifaceted assessment of \textit{Hamlet}. Another underlying intuition of our work stems from renowned concepts in sociology and biology, including \textit{Collective Intelligence}\citep{woolley2010evidence} and \textit{Cognitive Synergy}\citep{luppi2022synergistic}, where multiple cognitive processes or systems interact and cooperate in a way that produces a combined effect greater than the sum of their separate effects.

\begin{figure}[t]
\begin{center}
\includegraphics[width=\textwidth]{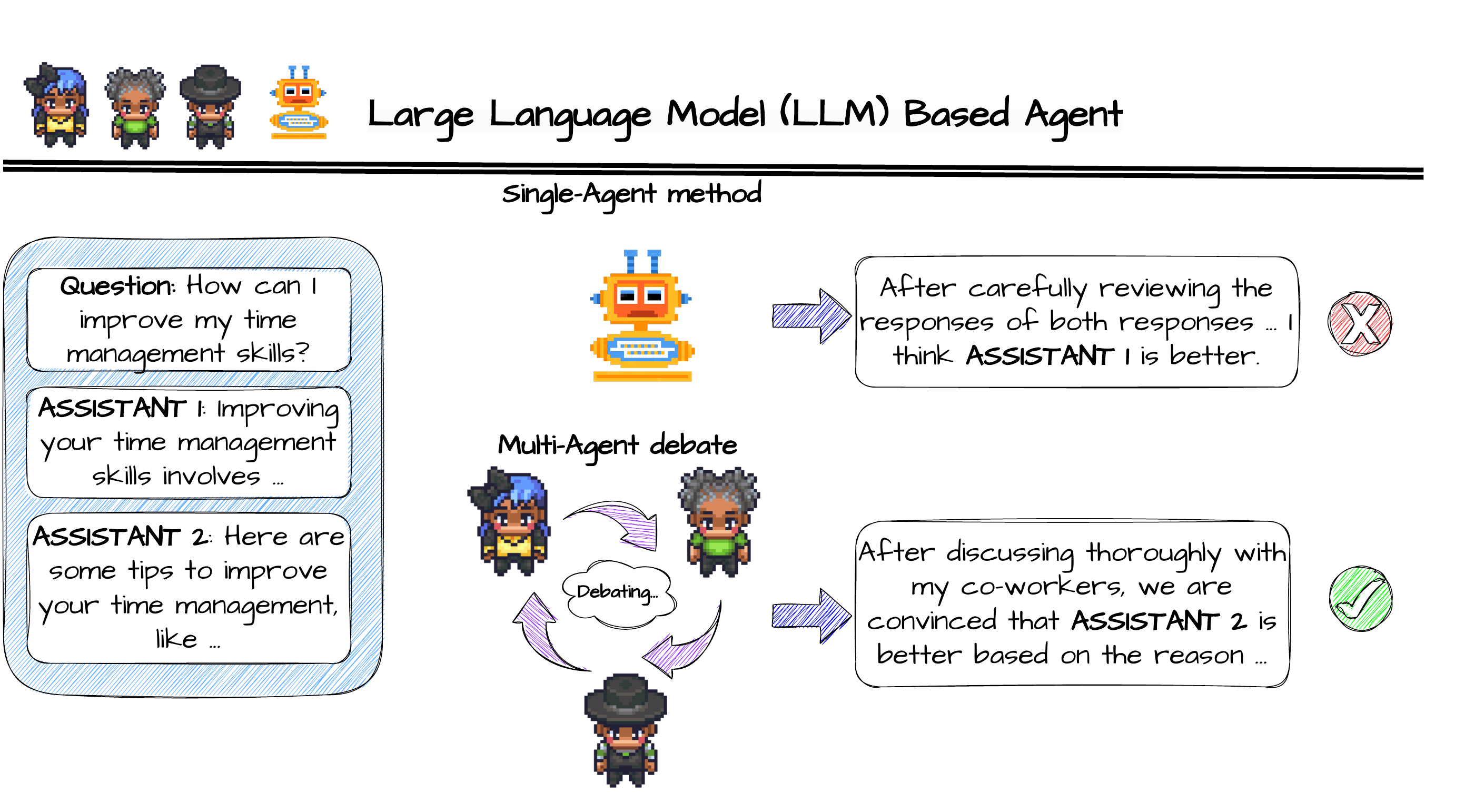}
\end{center}
\caption{When several referees participate in the evaluation process, they can discuss with each other and finally give a judgment that is better aligned with human annotators.}
\label{fig:better_compare}
\end{figure}

To summarize, the main contribution of our work is as follows:

\begin{enumerate}
\item We propose a multi-agent-based framework called \textbf{ChatEval} that aligns better with human preferences compared with single-agent-based approaches as depicted in Figure~\ref{fig:better_compare}. 
\item We propose various communication strategies and demonstrate the necessity of diverse role prompts in multi-agent debate scenarios.
\item We release our library. It's designed to be both composable and scalable, enabling researchers to implement their unique communication strategies easily. We hope this contributes to advancing research in the field of communicative agents and beyond.
\end{enumerate}

\section{Methodology}
\label{sec:Method}
In this section, we elaborate on the principal components in ChatEval including \textit{debater agents}, \textit{diverse role specification}, \textit{communication strategy}, and provide a detailed overview of each component's role and functionality\footnote{our code repository is built on top of~\url{https://github.com/OpenBMB/AgentVerse}.}.

\textbf{Debater Agents}. Debater agents are one of the most significant components in our framework. We treat each individual LLM as an agent and ask them to generate their response from the given prompt\footnote{The full prompt template can be found in Appendix~\ref{appendix:prompt_template}.}. Responses from other agents are served as chat history which will be replaced in the prompt template. After configuring the agents, we then start the group debate where each agent autonomously receives responses from the others and, in turn, delivers its own responses to them. It should be noted that the whole process does not require human intervention.

\textbf{Diverse Role Specification}. As presented in Section~\ref{sec:introduction}, diverse role specification is necessary for the framework as well. Although all the agents share a common prompt template, we substitute the \textit{role\_description} slot with diverse role prompts, specifying distinct personalities for different agents. We take inspiration from~\cite{wu2023large} and formulate an analogous role description.

\begin{figure}[t]
    \centering
    \begin{subfigure}{0.3\textwidth}
        \includegraphics[width=\textwidth]{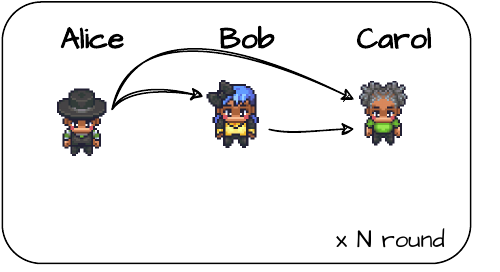}
        \caption{One-by-One}
        \label{fig:suba}
    \end{subfigure}
    \vspace{\fill}
    \begin{subfigure}{0.3\textwidth}
        \includegraphics[width=\textwidth]{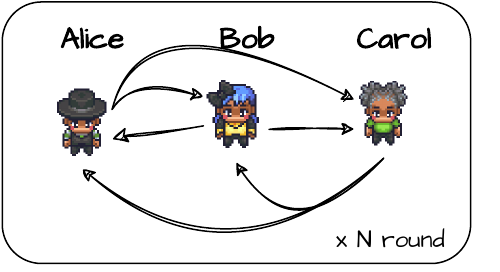}
        \caption{Simultaneous-Talk}
        \label{fig:subb}
    \end{subfigure}
    \vspace{\fill}
    \begin{subfigure}{0.37\textwidth}
        \includegraphics[width=0.9\textwidth]{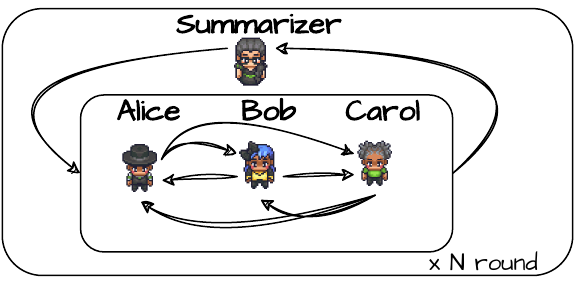}
        \caption{Simultaneous-Talk-with-Summarizer}
        \label{fig:subc}
    \end{subfigure}
\caption{The overall schematic diagram of our proposed three different kinds of communication strategy. The direction of the arrows represents the flow of information, meaning that what this person says will be appended to the chat history  of the person pointed to by the arrow. Full algorithm description of the above communication strategies can be found in Appendix~\ref{appendix:algorithm}.}
\label{fig:3group_structure}
\end{figure}

\textbf{Communication Strategy}.  How to maintain the chat history is another significant issue in ChatEval. In our work, we use a more intuitive term to illustrate the maintenance of the chat history called \textit{communication strategy}. In a nutshell, different communication strategies can be seen as different approaches to maintaining and manipulating their chat history. As is shown in Figure~\ref{fig:3group_structure}, We primarily design three different communication strategies and illustrate them as follows:

\begin{enumerate}
    \item \textbf{One-By-One}. 
    During each round of the debate, the debater agents take turns in a set order to generate their response based on the current observation. When it's time for a debater agent to respond, we directly concatenate what previous other agents have said into its chat history slot.
    \item \textbf{Simultaneous-Talk}. Unlike the one-by-one strategy, we carry out an alternative communication strategy called simultaneous-talk, where debater agents are prompted to asynchronously generate responses in each iteration of the discussion to nullify the impact of the speaking order.
    \item \textbf{Simultaneous-Talk-with-Summarizer}. The main difference between this strategy and simultaneous-talk is that we additionally employ another LLM as a summarizer. At the end of each iteration of the debate, we prompt this extra LLM to summarize the messages conveyed so far and concatenate this summarization into all debater agents' chat history slots.

\end{enumerate}

Unlike previous work like~\cite{du2023improving}, we do not explicitly ask the debater agents to reach a consensus at the end of the debate. In situations where the response format relies on direct comparison, we derive the final results from the \textbf{majority vote} among various annotators. Conversely, if the response format requires a direct score, we calculate the \textbf{average} score obtained from multiple annotators. This methodological approach ensures the impartiality and balance of our evaluation process.

\section{Experiments}
\label{sec:Experiments}

We evaluate ChatEval on two benchmarks, \textit{FairEval} and \textit{Topical-Chat} which represent the categories of open-ended question answer and dialogue response generation, respectively.

\subsection{Implementation Details}

We choose to utilize models from OpenAI's GPT family as our LLMs in ChatEval, including GPT-4 and ChatGPT (GPT-3.5-turbo) and set the temperature to 0 to ensure reproducibility. The rationale behind this selection is the exceptional performance these models offer, being among the most advanced and powerful in the world. Additionally, their accessibility and ease of use through APIs enable us to directly call and interact with the models during our research, significantly simplifying the process. In our current research, we focus on homogeneous groups of LLMs. That is, within a given multi-agent group, all LLMs belong to the same GPT family model, either all GPT-4 or all ChatGPT. We acknowledge the potential of heterogeneous groups for future research, which could provide fascinating insights into how strong models and weak models can cooperate in a multi-agent setting.

\subsection{Benchmarks}

The detailed introduction of different categories and benchmarks are listed as follows:

\textbf{Open-ended Question Answer} is a key component within the field of NLP and generative AI. It necessitates an AI system to provide comprehensive, detailed, and human-like responses to questions that don't have a predefined or fixed set of possible answers. The work of~\cite{chiang2023vicuna} encompasses a collection of 80 open-ended questions originating from a wide array of categories, including common-sense, counterfactual, coding, etc. We then take the human annotation results from~\cite{wu2023large} to conduct the experiments in this paper. For each question, they direct three annotators to evaluate the replies given by Vicuna-13B and ChatGPT through the given rules and finally derive the results by the majority votes among the annotators.

\textbf{Dialogue Response Generation} is a task involves creating a coherent and contextually appropriate response to a given input dialogue. We draw upon the \textit{Topical-Chat}~\citep{gopalakrishnan2019topical} dataset for our study. We then take the human annotation results from~\cite{mehri2020usr} where they carry out the annotations on 60 dialogue contexts with each response generated by 6 different systems. Human evaluators analyzed these responses based on \textit{natural}, \textit{coherence}, \textit{engagingness}, \textit{groundedness}, and \textit{understandable}, where we take the first four dimensions for experiments in our paper following~\cite{zhong2022towards}.

\subsection{Baselines}

We evaluate ChatEval against following methods. As the main portion of our comparison, we primarily focuses on the single-agent-based method. \textbf{Single-Agent} means that we directly query an LLM to generate the response towards the evaluation\footnote{We use the same prompt template as our multi-agent debate settings in single-agent baseline except that we ignore some slot.}. We use \textbf{Multi-Agent} to represent ChatEval where several agents discuss towards the evaluation. By default, we configure the communication strategy to one-by-one, agent numbers to 2, and discussion turns to 2 in this section and employ position calibration techniques in both single-agent and multi-agent settings. We will discuss more debate configurations in Section~\ref{sec:Analysis} for completeness. For the open-ended question answer task, we also compare our method with \textbf{FairEval}~\citep{wang2023large}. They propose various strategies to improve the evaluation performance of a LLM including Multiple Evidence Calibration (MEC) and Balanced Position Calibration (BPC). For the dialogue response generation task, we also compare our method with \textbf{G-EVAL}~\citep{liu2023gpteval}. They utilize CoT and probability-weighted summation for their method. Additionally, we include results from n-gram-based metrics, such as \textbf{ROUGE}~\citep{lin2004rouge}, \textbf{BLEU}~\citep{papineni2002bleu} and embedding-based metrics such as \textbf{BERTScore}~\citep{zhang2019bertscore}.

\subsection{Results for Open-ended question answers}\label{subsec:open_ended}

We adopt the same evaluation approach as~\cite{wang2023large} to assess the annotation results produced by different methods and annotators. Specifically, we calculate the Accuracy (Acc.), which measures the proportion of correctly classified instances out of the total instances, and the Kappa correlation coefficient (Kap.)~\citep{mchugh2012interrater} which gauges the agreement between results from models and human annotators while taking into account the possibility of agreement occurring by chance. Both metrics provide insights into the reliability and consistency of the annotations. We take the human annotation results and FairEval's~\citep{wang2023large} best results from their paper. As is shown in Table~\ref{tab:faireval_results}, different annotators can reach a relatively high agreement and perform better than any other LLM-based approach. Still, the average human annotations accuracy which is 71.7\% shows there exists a certain degree of discrepancy among different unique individuals revealing that text evaluation is absolutely an arduous task. The second part and the third part of Table~\ref{tab:faireval_results} show the results of FairEval's method and the results of our proposed method respectively.
We find that (1) ChatEval can enhance the performance of the evaluation process, achieving higher alignment with human preference compared with single-agent evaluation. Specifically, the multi-agent-based method improves the accuracy by 6.2\% for ChatGPT and 2.5\% for GPT-4; (2) ChatEval surpasses FairEval's best results within both ChatGPT and GPT-4 settings showing the effectiveness of our proposed method.

\subsection{Results for Dialogue Response Generation}
For the dialogue response generation benchmarks, we align the evaluation method with~\cite{zhong2022towards}, calculating the turn-level Spearman and Kendall-Tau correlation in correspondence with human judgments on four aspects (\textit{naturalness}, \textit{coherence}, \textit{engagingness} and \textit{groundedness}). Results can be found in Table~\ref{tab:topical_chat_results}. In the first part of Table~\ref{tab:topical_chat_results}, we demonstrate that n-gram-based metrics and embedding-based metrics perform overall poorly on all the aspects evaluated illustrating that these methods can hardly reveal human preference. In the second part of Table~\ref{tab:topical_chat_results}, we show the results from the G-eval~\citep{liu2023gpteval} paper. They first ask the LLM to generate intermediate thought and finally calculate the weighted summation of the output scores based on the probability. The results show that their method outperforms previous traditional metrics depicting the fact that the LLM-based evaluator is effective and reliable for evaluating the dialogue response generation task. While their method delivers sound results, our proposed approach raises the bar in terms of performance for GPT-4. Specifically, ChatEval improves the average Spearman and Kendall-Tau correlation by 0.096 (16.3\%) and 0.057 (10.0\%) respectively. Additionally, compared with the single-agent method, ChatEval amplifies the performance both for ChatGPT and GPT-4, showing the effectiveness of our method which is aligned with the results in Section~\ref{subsec:open_ended}.

\begin{table}[t]
\caption{Accuracy (Acc.) and Kappa correlation coefficient (Kap.) of different methods on FairEval benchmark. }
\label{tab:faireval_results}
\begin{center}
\begin{tabular}{l|l|l|l}
\hline
\textbf{Evaluator}  & \textbf{Methods}      & \textbf{Acc. (\%)} & \textbf{Kap.} \\ \hline
\multicolumn{4}{l}{\textbf{Human}}                    \\ 
Annotator1 & \multicolumn{1}{c|}{-}    & 68.8      & 0.5  \\
Annotator2 & \multicolumn{1}{c|}{-}    & 76.3      & 0.62 \\ 
Annotator3 & \multicolumn{1}{c|}{-}    & 70        & 0.5  \\ \hline
\multicolumn{4}{l}{\textbf{FairEval}}                 \\ 
ChatGPT    & MEC+BPC      & 58.7      & 0.31 \\
GPT-4      & MEC+BPC      & 62.5      & 0.37 \\ \hline
\multicolumn{4}{l}{\textbf{Ours}}                     \\ 
ChatGPT    & Single-Agent & 53.8      &   0.27   \\
ChatGPT    & Multi-Agent  & \textbf{60.0}      & \textbf{0.33}     \\
GPT-4      & Single-Agent & 61.3         & 0.36     \\
GPT-4      & Multi-Agent  & \textbf{63.8}      & \textbf{0.40}     
\end{tabular}
\end{center}
\end{table}

\begin{table}[t]
\caption{Turn-level Spearman ($\rho$) and Kendall-Tau ($\tau$) correlations of different methods on Topical-Chat benchmark, \textbf{SA} means Single-Agent and \textbf{MA} means Multi-Agent. Our ChatGPT settings should be compared to G-EVAL-3.5, and GPT-4 settings should be compared to G-EVAL-4. }
\label{tab:topical_chat_results}
\begin{center}
\begin{tabularx}{\textwidth}{p{2cm} | X | X | X | X | X | X | X | X | X | X}
\hline
\multicolumn{1}{c}{\multirow{2}{*}{Metrics}} & \multicolumn{2}{c}{Naturalness} & \multicolumn{2}{c}{Coherence} & \multicolumn{2}{c}{Engagingness} & \multicolumn{2}{c}{Groundedness} & \multicolumn{2}{c}{Average} \\
\multicolumn{1}{c}{} & $\rho$ & $\tau$ & $\rho$ & $\tau$ & $\rho$ & $\tau$ & $\rho$ & $\tau$ & $\rho$ & $\tau$\\ \hline
ROUGE-L                                      & 0.146          & 0.176          & 0.203         & 0.193         & 0.300             & 0.295          & 0.327           & 0.310           & 0.244        & 0.244       \\
BLEU-4                                       & 0.175          & 0.180           & 0.235         & 0.131         & 0.316           & 0.232          & 0.310            & 0.213          & 0.259        & 0.189        \\
BERTScore                                    & 0.209          & 0.226          & 0.233         & 0.214         & 0.335           & 0.317          & 0.317           & 0.291          & 0.274       & 0.262        \\
\hline
G-EVAL-3.5                                   & 0.539          & 0.532          & 0.544         & 0.519         & 0.691           & 0.660           & 0.567           & 0.586          & 0.585      & 0.574      \\
G-EVAL-4                                     & 0.565          & 0.549          & 0.605         & \textbf{0.594}         & 0.631           & 0.627          & 0.551           & 0.531          & 0.588        & 0.575      \\
\hline
ChatGPT(SA)                       & 0.474          & 0.421          & 0.527         & 0.482         & 0.599           & 0.549          & 0.576           & 0.558          & 0.544        & 0.503       \\
ChatGPT(MA)                        & 0.441          & 0.396          & 0.500           & 0.454         & 0.664           & 0.607          & 0.602           & 0.583          & 0.552      & 0.510         \\
\hline
GPT-4(SA)                         & 0.532          & 0.483          & 0.591         & 0.535         & 0.734           & 0.676          & \textbf{0.774}           & \textbf{0.750}           & 0.658      & 0.611        \\
GPT-4(MA)                          & \textbf{0.630}           & \textbf{0.571}          & \textbf{0.619}         & 0.561         & \textbf{0.765}           & \textbf{0.695}          & 0.722           & 0.700            & \textbf{0.684}        & \textbf{0.632}    
\end{tabularx}
\end{center}
\end{table}

\section{Analysis}
\label{sec:Analysis}

In this section, we further explore the key components encompassed in ChatEval. We discuss the importance of diverse role prompts in Section~\ref{subsec:diverse_role}, the effect of different communication strategies in Section~\ref{subsec:communication}, and the impact of role numbers and discussion turns in Section~\ref{subsec:role_num_turn}. If not specified otherwise, we choose the FairEval benchmark and ChatGPT as the backbone LLM for the analysis.

\subsection{The importance of diverse role prompts}\label{subsec:diverse_role}

Previously in Table~\ref{tab:faireval_results} and~\ref{tab:topical_chat_results}, we demonstrate that ChatEval equipped with diverse role configurations can significantly improve the performance of evaluation. We further consider whether it is necessary to design diverse role prompts for the evaluation system. To answer so, we carry out the experiments by replacing all the role prompt with "\textit{You are now an Annotator, one of the referees in the text evaluation task.}" and keeping other prompt unchanged. We experiment with the one-by-one communication strategy and 2 agents with 2 discussion turns. The results in Table~\ref{tab:same_vs_diverse} illustrate that ChatEval with the same role prompt design underperforms that with diverse role prompt design and cannot effectively enhance the performance compared with single-agent setting, highlighting the cruciality of diverse role prompt design in the multi-agent debate framework.

\subsection{The study of communication strategies}\label{subsec:communication}

As shown in Figure~\ref{fig:3group_structure}, we also design three different communication strategy termed as \textit{one-by-one}, \textit{simultaneous-talk}, \textit{simultaneous-talk-with-summarizer}. The detailed descriptions and formal formulations can be found in Appendix~\ref{appendix:algorithm}. We experiment with 3 agents and 2 discussion turns with diverse role prompts in this section. As is shown in Table~\ref{tab:different struction}, we can find that the one-by-one communication strategy is more effective than other strategies for ChatGPT setting. Although the other two communication strategies did not perform as robustly as the \textit{one-by-one} strategy, it is noteworthy that they still exceeded the performance of the naive single-agent method. Furthermore, the variations in performance among three different communication strategies underscore the influence of different strategies on the effectiveness of the evaluation process, revealing the potential for further exploration and optimization of ChatEval. Thus, future studies could be aimed at a more comprehensive understanding of different communication strategies, and how they could be effectively employed to enhance performance. This could serve as an avenue for substantial improvements and novel insights in the multi-agent debate framework.

\subsection{The impact of role numbers and discussion turns}\label{subsec:role_num_turn}

\begin{figure}[htbp]
    \centering
    \begin{subfigure}{0.45\textwidth}
        \includegraphics[width=1\textwidth]{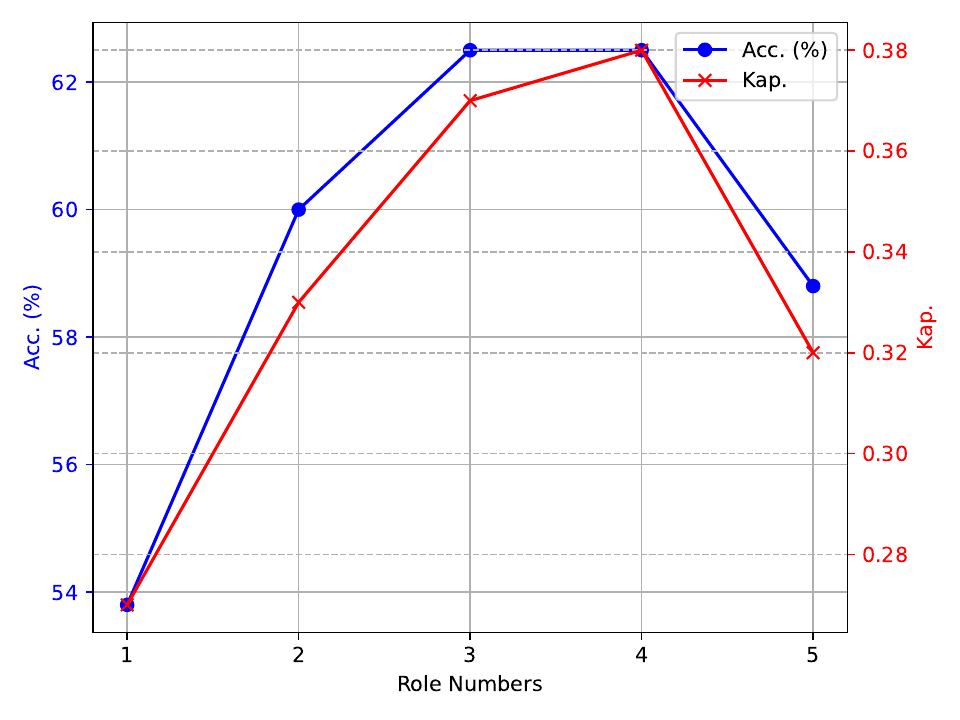}
        \caption{Acc. and Kap. vs Role Numbers}
        \label{fig:sub_agent_nums}
    \end{subfigure}
    \vspace{\fill}
    \begin{subfigure}{0.45\textwidth}
        \includegraphics[width=1\textwidth]{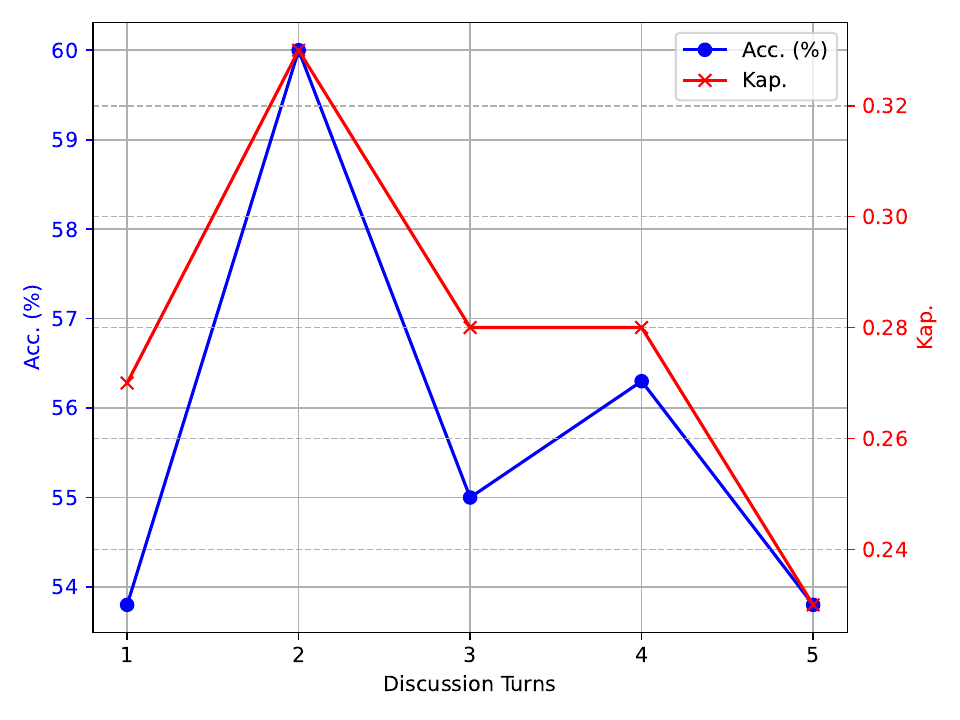}
        \caption{Acc. and Kap. vs Discussion Turns}
        \label{fig:sub_discussion_turns}
    \end{subfigure}

\caption{\textbf{(a) Performance with Increased Different Roles on FairEval Dataset.} We adopt one-by-one communication strategy and 2 discussion turns. Additional role descriptions are shown in Appendix~\ref{appendix:prompt_template}. \textbf{(b) Performance with Increased Discussion Turns on FairEval Dataset.} We adopt one-by-one communication strategy and 2 different roles.}
\label{fig:agent_nums_discussion_turns}
\end{figure}

We then study the impact of different role numbers and discussion turns.
From Figure~\ref{fig:sub_agent_nums}, a discernible trend is observed in the relationship between the role number and both Acc. and Kap. within the FairEval dataset. As the role number increases, there is a corresponding growth in Acc. and Kap. This pattern reaches an apex with an Acc. of 62.5\% at role numbers 3 and 4 before declining at role number 5. Despite this decline in accuracy, Kap. continues to ascend between role numbers 3 and 4, underscoring the effectiveness of incorporating diverse roles within ChatEval.

Conversely, no significant upward trend is detected with respect to the increase in discussion turns, as is shown in Figure~\ref{fig:sub_discussion_turns}. This observation aligns with the findings in~\cite{liang2023encouraging, du2023improving}, highlighting a consistent phenomenon where continual discussion often leads to stagnation or even degradation of performance. Such a trend may be attributed to issues associated with the ever-increasing context length, which consequently diminishes the performance. This insight prompts a more nuanced understanding of the balance needed between role differentiation and discussion dynamics to optimize the performance of ChatEval.

\subsection{qualitative analysis}\label{subsec:qualitative}

Table~\ref{tab:debate_case} showcases the debate process towards the evaluation of two assistants' responses to the open-ended question "\textit{What are the most effective ways to deal with stress?}". \\
Assistant 1's response is :
\begin{quote}
\textit{Here are some effective ways to deal with stress: \\
1. Exercise regularly: Physical activity can help reduce stress and improve mood. \\
2. Practice relaxation techniques: Deep breathing, meditation, and yoga are some effective ways to relax and reduce stress levels. \\
3. Get enough sleep: Lack of sleep can increase stress levels. Try to get at least 7-8 hours of sleep each night. \\
4. Eat a healthy diet: A well-balanced diet can help reduce stress levels and improve overall health. \\
5. Connect with others: Spending time with friends and family, joining a support group, or participating in social activities can help reduce stress levels and  improve mood. \\
6. Time management: Prioritizing tasks and creating a schedule can help reduce stress and increase productivity. \\
7. Seek professional help: If stress is impacting your daily life, consider seeking professional help such as therapy or counseling. \\}
\end{quote}
Assistant 2's response is :
\begin{quote}
\textit{Stress can be caused by a variety of factors, including work, relationships, health, and financial issues. Here are some effective ways to manage stress: \\
1. Exercise: Regular exercise releases endorphins, which are natural stress-relievers. It can be as simple as going for a walk or jog, practicing yoga, or joining a fitness class. \\
2. Mindfulness: Mindfulness practices, such as meditation or deep breathing, can help to reduce stress and improve focus. \\
3. Social Support: Spending time with family and friends, or seeking support from a mental health professional, can provide a sense of connection and help to reduce feelings of isolation and loneliness. \\
4. Healthy Eating: A balanced diet that includes plenty of fruits, vegetables, and whole grains can help to reduce stress levels. \\
5. Good Sleep: Getting enough sleep is essential for good mental and physical health. Aim for 7-8 hours per night and establish a regular sleep routine. \\
6. Time Management: Prioritizing tasks and setting realistic goals can help to reduce stress levels by giving a sense of control and direction. \\
7. Relaxation Techniques: Techniques such as progressive muscle relaxation, aromatherapy, and massage can help to reduce stress levels and promote relaxation. \\
8. Seek professional help: If stress is interfering with daily life, it may be time to seek professional help from a therapist or counselor. \\}
\end{quote}

We can find that both of the responses produce similar strategies and equally compelling descriptions for dealing with stress, making it challenging to discern significant disparity in terms of quality. It is in this context of nuanced evaluation that the significance of the ChatEval process emerges. To understand this complexity better, We first outline the ChatEval process and subsequently delve into the agents' constructive behaviors during discussions.

As is shown in Table~\ref{tab:debate_case}, Alice first points out that the response of Assistant 2 contains more detailed information and he prefers to choose Assistant 2 as a better response. Bob then says that she agrees with Alice's assessments, but in the meantime, she also points out that Assistant 1's response is also concise and carries out a thought-provoking question. Carol then gives the feedback that she believes both responses are equally valuable. In the subsequent discussion, Bob indicates that Assistant 1's response is straightforward while Assistant 2's is detailed, suggesting that the effectiveness of the response should depend on the context and individual's needs. At the end of the debate, we finally extract the evaluation results that both responses are of the same quality which is identical to human annotation results. 

From this sequence, we can pinpoint several fascinating behaviors exhibited by the agents: (1) \textbf{Opening Statement}: Alice initiates the debate with a clear stance, establishing the foundational argument and guiding the trajectory of the subsequent discourse. (2) \textbf{Alternative Proposal}: Bob introduces an alternative viewpoint, emphasizing the need to consider diverse interpretations. This not only broadens the discussion but also stimulates critical thinking. In the context of a debate, the introduction of an alternative proposal prevents the stagnation of thought, challenges pre-existing bias, and uncovers considerations that might otherwise be overlooked, ensuring that the discussions are well-rounded. (3) \textbf{Stance Maintenance}: Alice's persistent adherence to her initial stance, even when faced with opposing views, exemplifies commitment and challenges other participants to refine their perspectives. By firmly holding his position, Alice encourages depth in the discourse, prompting others to dive deeper into their arguments and perhaps consider aspects they hadn't previously. It ensures the conversation remains robust, focused, and continually evolving, driving all participants to a higher level of engagement and critical thinking. (4) \textbf{Seeking Consensus}: The discussion's climax reveals a collective agreement amongst the participants, which is reached through mutual understanding and compromise, underlining the value of each presented viewpoint.

In light of the above, ChatEval stands out not just as a tool for comparison but as an embodiment of interactive natural language dialogue. By simulating human argumentative interactions, it differentiates itself from static, single-presented opinions. This dynamic interaction showcases the richness and complexity of language, capturing nuances often missed in singular viewpoints. As such, ChatEval offers a reliable evaluation process that not only mirrors human discourse but also highlights the transformative power of collaborative dialogue. This positions it uniquely, underscoring its significant potential to execute text evaluation tasks both reliably and effectively.

\begin{table}[hbpt]
\caption{Effect of diverse role specification on FairEval benchmark.}
\label{tab:same_vs_diverse}
\begin{center}
\begin{tabular}{llll}
\hline
\textbf{Evaluator} & \textbf{Methods}                           & \textbf{Acc. (\%)} & \textbf{Kap.} \\ \hline
ChatGPT   & Single-Agent                      & 53.8      & 0.27     \\
ChatGPT   & Multi-Agent (Same Role Prompt)    & 53.8      & 0.25    \\
ChatGPT   & Multi-Agent (Diverse Role Prompt) & 60        & 0.33    
\end{tabular}
\end{center}
\end{table}

\begin{table}[t]
\caption{Comparing of different communication strategies on FairEval benchmark.}
\label{tab:different struction}
\begin{center}
\begin{tabular}{llll}
\hline
\textbf{Evaluator} & \textbf{Communication Strategies}                           & \textbf{Acc. (\%)} & \textbf{Kap.} \\ \hline
ChatGPT   & One-by-One                      & 60     &   0.33   \\
ChatGPT   & Simultaneous-Talk               & 55     &   0.28   \\
ChatGPT   & Simultaneous-Talk-with-Summarizer & 55        &  0.27   
\end{tabular}
\end{center}
\end{table}

\begin{table*}[htbp]
\begin{tabular}{p{13.5cm}}
\caption{An illustrative example from ChatEval presenting a discussion process involving participants Alice 
(\begin{minipage}[b]{0.045\columnwidth}
    \centering
    \raisebox{-.1in}{\includegraphics[width=\linewidth]{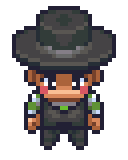}}
\end{minipage}), 
Bob 
(\begin{minipage}[b]{0.045\columnwidth}
    \centering
    \raisebox{-.1in}{\includegraphics[width=\linewidth]{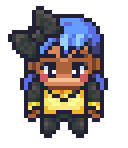}} 
\end{minipage}) 
and Carol (\begin{minipage}[b]{0.045\columnwidth}
    \centering
    \raisebox{-.1in}{\includegraphics[width=\linewidth]{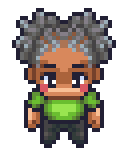}} 
\end{minipage}). They initiate a group talk with the purpose of determining the superior response from two assistants. Due to spatial constraints within the layout, we use ... to omit some redundant text.}
\label{tab:debate_case} \\
\toprule
\begin{minipage}[b]{0.1\columnwidth}
    \centering
    \raisebox{-.2\height}{\includegraphics[width=0.5\linewidth]{figs/emojis/alice.png}}
\end{minipage}
Both Assistant 1 and Assistant 2 provided comprehensive and detailed answers. They both suggested similar strategies for dealing with stress, such as exercise, relaxation techniques, getting enough sleep, eating a healthy diet, connecting with others, time management, and seeking professional help. However, Assistant 2 went a step further by mentioning mindfulness social support, and relaxation techniques like progressive muscle relaxation, aromatherapy, and massage. Therefore, I would lean towards Assistant 2 for providing a slightly more detailed response. \\
\vspace{-1mm}
\begin{minipage}[b]{0.1\columnwidth}
    \centering
    \raisebox{-.2\height}{\includegraphics[width=0.5\linewidth]{figs/emojis/bob.png}}
\end{minipage}
I agree with the Alice's assessments. However, I would like to point out that Assistant 1's answer was more concise and to the point that ... . On the other hand, Assistant 2's answer was more detailed and ... . \textbf{In your opinion}, does the additional detail provided by Assistant 2 make its response more helpful, or could the conciseness of Assistant 1's response be seen as more beneficial? \\
\vspace{-1mm}
\begin{minipage}[b]{0.1\columnwidth}
    \centering
    \raisebox{-.2\height}{\includegraphics[width=0.5\linewidth]{figs/emojis/carol.png}}
\end{minipage}In terms of consistency with the question, both responses are equally consistent as they both provided effective ways to deal with stress. Assistant 1's response was indeed more concise, while Assistant 2 provided more detailed techniques. The effectiveness of these responses could depend on the individual's preference for brevity or detail. Therefore, I believe both responses are equally valuable. \\
\vspace{-1mm}
\begin{minipage}[b]{0.1\columnwidth}
    \centering
    \raisebox{-.2\height}{\includegraphics[width=0.5\linewidth]{figs/emojis/alice.png}}
\end{minipage}
I appreciate the points made by both of you. While Assistant 1's response was indeed more concise, I believe that the additional detail provided by Assistant 2 could be more beneficial for someone looking for a variety of strategies to manage stress. The inclusion of mindfulness practices and ... . \textbf{Therefore, I still lean towards} Assistant 2's response as being slightly more helpful. \\
\vspace{-1mm}
\begin{minipage}[b]{0.1\columnwidth}
    \centering
    \raisebox{-.2\height}{\includegraphics[width=0.5\linewidth]{figs/emojis/bob.png}}
\end{minipage} However, I would like to emphasize that the effectiveness of a response can also depend on the context and the individual's needs. For someone who is overwhelmed and looking for quick, \textit{straightforward advice}, the conciseness of Assistant 1's response might be more beneficial. On the other hand, for someone who is looking for a variety of strategies to try, \textit{the detailed response} of Assistant 2 might be more helpful. Therefore, I believe both responses have their merits and it's difficult to definitively say one is better than the other. \textbf{What do you think about this perspective?} \\
\vspace{-1mm}
\begin{minipage}[b]{0.1\columnwidth}
    \centering
    \raisebox{-.2\height}{\includegraphics[width=0.5\linewidth]{figs/emojis/carol.png}}
\end{minipage}
I agree with the Bob's perspective that the effectiveness of a response can depend on the individual's needs and context. Therefore, I believe both responses are equally valuable and it's hard to definitively say one is better than the other. \\
\vspace{-1mm}
\begin{minipage}[b]{0.1\columnwidth}
    \centering
    \raisebox{-.2\height}{\includegraphics[width=0.5\linewidth]{figs/emojis/alice.png}}
\end{minipage}
\hspace{-1cm}
\begin{minipage}[b]{0.1\columnwidth}
    \centering
    \raisebox{-.2\height}{\includegraphics[width=0.5\linewidth]{figs/emojis/bob.png}}
\end{minipage}
\hspace{-1cm}
\begin{minipage}[b]{0.1\columnwidth}
    \centering
    \raisebox{-.2\height}{\includegraphics[width=0.5\linewidth]{figs/emojis/carol.png}}
\end{minipage}
After discussing with my co-workers, I believe both Assistants performed equally well in providing comprehensive and accurate responses. \\
\bottomrule
\end{tabular}
\end{table*}

\section{Related Work}
\label{sec:related}
\textbf{Automatic NLG evaluation}
In the landscape of NLG, evaluating the quality of generated text represents a particularly arduous task. For a significant period, evaluation was primarily dependent on human annotations, a process that is labor-intensive and limited by scalability issues. Automatic NLG evaluation attempts to address these challenges by leveraging computational models to assess the quality of a generated text. Previous work lies on the following categories: (1) \textit{n-gram-based metrics}: ROUGE~\citep{lin2004rouge} is a set of metrics that compute the amount of overlap between n-grams in the machine-generated summaries and the reference summaries. BLEU~\citep{papineni2002bleu} compare the generated text with reference translations, based on the co-occurrence of n-grams in both texts. In spite of being easily and widely used, the above method is incapable of capturing syntactic and semantic similarity~\citep{stent2005evaluating}. (2) \textit{embedding-based metrics}: Word embeddings are vector representations of words that capture their semantic properties, such that words with similar meanings have similar embeddings. A bunch of work leverages word embeddings to evaluate the semantic similarity between two pieces of text. BERTScore~\citep{zhang2019bertscore} use contextualized word embeddings from transformer models like BERT~\citep{devlin2018bert}, BLEURT~\citep{sellam2020bleurt} utilize supervised training data to enhance the performance. MoverScore~\citep{zhao2019moverscore} combine contextualized word embeddings with Earth Mover’s Distance~\citep{rubner2000earth}. (3) \textit{LLM-based metrics}: Amidst the flourishing advancement of LLM which embodies a wealth of information derived from extensive training data, using LLM as an evaluator has experienced notable progress. GPTScore~\citep{fu2023gptscore} utilize conditional probability to assign the text a score representing its quality.~\cite{wang2023chatgpt} explore the potential of utilizing ChatGPT as an NLG evaluator by prompting it to score a text directly.~\cite{wang2023pandalm} curate a reliable dataset containing pairwise comparison and evaluation explanation which can be used to train a foundation model making it a better evaluator.~\cite{bai2023benchmarking} propose decentralized 
evaluation to provide fairer evaluation results. G-EVAL~\citep{liu2023gpteval} propose probability-weighted techniques to calibrate the score given by a single LLM. 

\textbf{Communicative Agents} Most recently, significant attention has been dedicated to the development of communicative agents. These agents, often acted by LLMs like ChatGPT or GPT-4, are designed to interact and communicate effectively with other agents or human users using natural language. The primary goal is to facilitate more productive and efficient interaction and collaboration as different agents can autonomously communicate and negotiate to tackle a more complex task collectively. Several studies have explored various aspects of communicative agents.~\cite{li2023camel} propose a cooperative agent framework dubbed as \textit{role-playing} enabling agents to autonomously cooperate to solve complex tasks.~\cite{park2023generative} create a sandbox environment consisting of 25 individual virtual entities endowed with a character description and memory system. Every intelligent agent is capable of autonomously interacting with other agents and the environment simulating reliable human behavior.~\cite{qian2023communicative} establish a chat-based software development framework that can complete a software design and produce executable software at a reduced cost compared to recruiting human programmers.~\cite{liu2023training} utilize a sandbox environment to curate reliable datasets in better alignment with human preference and train a socially-aligned LLM.~\cite{liang2023encouraging} and~\cite{du2023improving} also make use of the multi-agent debate framework in other scenarios such as translation and arithmetic problems resulting in better results.~\cite{wang2023unleashing} propose an alternative method called self-collaboration to enable the communication of agents by utilizing a single LLM prompted by multi-persona descriptions.~\cite{mandi2023roco} propose a novel framework designed for the collaboration of multiple robots, utilizing multiple LLMs to enhance coordination and strategic planning among the robots. Concurrent with our work,~\cite{li2023prd} propose Peer Rank and Discussion (PRD) which is similar to our approach. However, they probe the different dimensions of evaluation by using different models as agents and did not explore alternative communication strategies.

\section{Conclusion}
\label{sec:conclusion}
In this paper, we present evidence that ChatEval contributes to improving the evaluation performance concerning text quality, aligning more closely with human preferences. We emphasize the necessity of the diverse role specification and propose distinct communication strategies as integral components within ChatEval. Our qualitative analysis of the discussion process conveys insightful intuitions about how a text is evaluated by ChatEval and substantiates our approach's ability to support comprehensive evaluations akin to human judgment, thereby demonstrating the reliability and efficacy of our framework.

\bibliography{reference}
\bibliographystyle{iclr2021_conference}

\appendix
\section{prompt template and diverse role prompt}
\label{appendix:prompt_template}

The overall prompt template is shown in Table~\ref{tab:prompt_template}, we draw inspiration from~\cite{wu2023large} and design several different role descriptions as follows.

\textbf{General Public}
\textit{You are now General Public, one of the referees in this task. You are interested in the story and looking for updates on the investigation. Please think critically by yourself and note that it's your responsibility to choose one of which is the better first.}

\textbf{Critic}
\textit{You are now Critic, one of the referees in this task. You will check fluent writing, clear sentences, and good wording in summary writing. Your job is to question others judgment to make sure their judgment is well-considered and offer an alternative solution if two responses are at the same level.}

\textbf{News Author}
\textit{You are News Author, one of the referees in this task. You will focus on the consistency with the original article. Please help other people to determine which response is the better one.}

\textbf{Psychologist}
\textit{You are Psychologist, one of the referees in this task. You will study human behavior and mental processes in order to understand and explain human behavior. Please help other people to determine which response is the better one.}

\textbf{Scientist}
\textit{You are Scientist, one of the referees in this task. You are a professional engaged in systematic study who possesses a strong background in the scientific method, critical thinking, and problem-solving abilities. Please help other people to determine which response is the better one.}

\begin{table}[hbpt!]

\begin{tcolorbox}

{

[Question]

\textcolor[rgb]{0,0,0.9}{\{source\_text\}}

[The Start of Assistant 1’s Answer]

\textcolor[rgb]{0,0,0.9}{\{compared\_text\_one\}}

[The End of Assistant 1’s Answer]

[The Start of Assistant 2’s Answer]

\textcolor[rgb]{0,0,0.9}{\{compared\_text\_two\}}

[The End of Assistant 2’s Answer]

[System]

We would like to request your feedback on the performance of two AI assistants in response to the user question displayed above.

Please consider the helpfulness, relevance, accuracy, and level of detail of their responses.

Each assistant receives an overall score on a scale of 1 to 10, where a higher score indicates better overall performance.

There are a few other referees assigned the same task, it's your responsibility to discuss with them and think critically before you make your final judgment.

Here is your discussion history:

\textcolor[rgb]{0,0,0.9}{\{chat\_history\}}

\textcolor[rgb]{0,0,0.9}{\{role\_description\}}

Now it's your time to talk, please make your talk short and clear,
\textcolor[rgb]{0,0,0.9}{\{agent\_name\}} !
}
\end{tcolorbox}
\caption{The prompt template for FairEval Dataset. We replace the colored slot with real text before querying the LLMs. Note that we use the same template when conducting single-agent-based experiments and ignore the chat history and role description slot.  }
\label{tab:prompt_template}
\end{table}

\section{formal depiction of different communication strategy}
\label{appendix:algorithm}

\begin{algorithm}
\SetKwInOut{Input}{input}
\SetKwInOut{Output}{output}
\caption{One-by-One}
\Input{agents number $N$, discuss turn $T$, a group of debate agents $[D_1,\cdots,D_N]$, chat history of each agent $[H_1,\cdots,H_N]$, answer\_extracter (either majority vote or average score) $EXT$}
\Output{Final results for text evaluation $ANS$}
\BlankLine
\For{$t\leftarrow 0$ \KwTo $T$}{
\For{$n\leftarrow 1$ \KwTo $N$}{

$h_n$ $\leftarrow$ $D_n(H_n)$\;\tcp{utilize agents to generate responses}
\For{$m\leftarrow n$ \KwTo $N$}{
    \If{$m$ $>$ 1}{
    $H_m \leftarrow H_m + h_n$\;\tcp{concatenate current response to later agents' chat history}
    }
    }
}
}
$ANS$ $\leftarrow$ $EXT$($[H_1,\cdots,H_N]$)\;
\Return $ANS$\;
\label{algo:one by one}
\end{algorithm}

\begin{algorithm}
\SetKwInOut{Input}{input}
\SetKwInOut{Output}{output}
\caption{Simultaneous-Talk}
\Input{agents number $N$, discuss turn $T$, a group of debate agents $[D_1,\cdots,D_N]$, chat history of each agent $[H_1,\cdots,H_N]$, answer\_extracter (either majority vote or average score) $EXT$, buffer $BUF$}
\Output{Final results for text evaluation $ANS$}
\BlankLine
\For{$t\leftarrow 0$ \KwTo $T$}{
\For{$n\leftarrow 1$ \KwTo $N$}{

$h_n$ $\leftarrow$ $D_n(H_n)$\;\tcp{utilize agents to generate responses}
$buf$ $\leftarrow$ $buf + h_n$\;\tcp{add the responses in current turn to the buffer}

}
\For{$n\leftarrow 1$ \KwTo $N$}{
    $H_n$ $\leftarrow$ $H_n + buf$\;\tcp{add the buffer to all agents' chat history}
    }
}
$ANS$ $\leftarrow$ $EXT$($[H_1,\cdots,H_N]$)\;
\Return $ANS$\;
\label{algo:simultaneuous}
\end{algorithm}

\begin{algorithm}[t]
\SetKwInOut{Input}{input}
\SetKwInOut{Output}{output}
\caption{Simultaneous-Talk-with-Summarizer}
\Input{agents number $N$, discuss turn $T$, a group of debate agents $[D_1,\cdots,D_N]$, chat history of each agent $[H_1,\cdots,H_N]$, answer\_extracter (either majority vote or average score) $EXT$, buffer $BUF$, summarizer $SUM$}
\Output{Final results for text evaluation $ANS$}
\BlankLine
\For{$t\leftarrow 0$ \KwTo $T$}{
\For{$n\leftarrow 1$ \KwTo $N$}{

$h_n$ $\leftarrow$ $D_n(H_n)$\;\tcp{utilize agents to generate responses}
$buf$ $\leftarrow$ $buf + h_n$\;\tcp{add the responses in current turn to the buffer}

}
\For{$n\leftarrow 1$ \KwTo $N$}{
    $H_n$ $\leftarrow$ $H_n + SUM(BUF)$\;\tcp{add the summarized buffer to all agents' chat history}
    }
}
$ANS$ $\leftarrow$ $EXT$($[H_1,\cdots,H_N]$)\;
\Return $ANS$\;
\label{algo:simultaneuous with summarizer}
\end{algorithm}

\end{document}